\documentclass[11pt]{article}
\usepackage[preprint]{acl}
\usepackage{times}
\usepackage{latexsym}
\usepackage{algorithm}
\usepackage[T1]{fontenc}
\usepackage[utf8]{inputenc}
\usepackage{microtype}
\usepackage{inconsolata}
\usepackage{graphicx}
\usepackage{tcolorbox}
\usepackage{multicol}
\usepackage{url}
\usepackage[table]{xcolor}
\definecolor{lightgray}{gray}{0.9}
\usepackage{amsmath}
\usepackage{listings}
\usepackage{xcolor}
\definecolor{dkgreen}{rgb}{0,0.6,0}
\definecolor{gray}{rgb}{0.5,0.5,0.5}
\definecolor{mauve}{rgb}{0.58,0,0.82}
\title{Interleaved Tool-Call Reasoning for Protein Function Understanding}

\author{
 \textbf{Chuanliu Fan\textsuperscript{1}},
 \textbf{Zicheng Ma\textsuperscript{3}}, \textbf{Huanran Meng\textsuperscript{1}},
 \textbf{Aijia Zhang\textsuperscript{1}},
\\
 \textbf{Wenjie Du\textsuperscript{4}},
 \textbf{Jun Zhang\textsuperscript{3}},
 \textbf{Yi Qin Gao\textsuperscript{3}},
 \textbf{Ziqiang Cao\textsuperscript{1,2}},
 \textbf{Guohong Fu\textsuperscript{1,2}},
\\
 \textsuperscript{1}School of Computer Science and Technology, Soochow University,\\
 \textsuperscript{2}Institute of Artificial Intelligence, Soochow University,\\
 \textsuperscript{3}Changping Laboratory,\\
 \textsuperscript{4}School of Software Engineering, USTC.
}

\begin{document}
\maketitle
\begin{abstract}
Recent advances in large language models (LLMs) have highlighted the effectiveness of chain-of-thought reasoning in symbolic domains such as mathematics and programming. However, our study shows that directly transferring such text-based reasoning paradigms to protein function understanding is ineffective: reinforcement learning mainly amplifies superficial keyword patterns while failing to introduce new biological knowledge, resulting in limited generalization. We argue that protein function prediction is a knowledge-intensive scientific task that fundamentally relies on external biological priors and computational tools rather than purely internal reasoning.
To address this gap, we propose PFUA, a tool-augmented protein reasoning agent that unifies problem decomposition, tool invocation, and grounded answer generation. Instead of relying on long unconstrained reasoning traces, PFUA integrates domain-specific tools to produce verifiable intermediate evidence. Experiments on four benchmarks demonstrate that PFUA consistently outperforms text-only reasoning models with an average performance improvement of $103\%$\footnote{\url{https://github.com/better-fcl/pfua}}.
\end{abstract}

\section{Introduction}
Understanding protein function is a fundamental task in computational biology with broad implications in drug discovery, disease understanding, and synthetic biology.
Despite the exponential growth in protein sequence databases, a significant portion of proteins lack reliable functional annotations~\cite{Zhou653105}.
Wet-lab experiments to determine protein function are time-consuming and resource-intensive, motivating scalable computational pipelines for automated functional annotation from sequence information.
The task of protein function understanding aims to automatically predict the biological roles of proteins using computational models~\cite{fang2024molinstructions}.
These include predictions of catalytic reactions, cellular functions based on Gene Ontology (GO) terms, and the identification of conserved domains or sequence motifs.
Traditional supervised fine-tuning (SFT) approaches learn a direct mapping from protein sequences to functional outputs, often achieving competitive performance.
These models encode protein modality features effectively, but their predictions remain largely uninterpretable.
As a result, understanding the reasoning process behind functional predictions, and enabling models to generalize beyond pattern matching remains a persistent challenge.

Recently, the DeepSeek R1 model has demonstrated remarkable reasoning capabilities enabled by cold-start reasoning data construction and GRPO-based reinforcement learning.
These methods significantly enhance reasoning performance in symbolic domains such as mathematics and code generation.
Inspired by R1-style text-based reasoning, we construct a cold-start protein reasoning dataset using \emph{kimi-k2-0905-preview}.
We then train the Qwen2.5-3B model with SFT followed by reinforcement learning, using a mixture of format rewards and accuracy rewards computed via ROUGE\_L and F1.
However, our early explorations reveal a notable discrepancy between protein reasoning tasks and symbolic reasoning tasks.
Without SFT on Kimi-generated reasoning data, the Qwen2.5-3B-R1-Zero model primarily receives format rewards while failing to achieve sufficient accuracy rewards.
After cold-start supervised fine-tuning, the initial reward increases substantially, but subsequent improvement plateaus quickly, converging around $0.4$.
Inspection of the model’s generated rationales shows that the model tends to rely on repetitive high-frequency keywords to accumulate partial rewards, rather than identifying biologically meaningful functional cues.
This failure mode highlights the science knowledge-intensive nature of protein function understanding: unlike mathematics or programming, protein-related questions cannot be solved through symbolic reasoning alone.
This finding aligns with conclusions from prior work~\cite{yue2025does}, which shows that reinforcement learning mainly improves the sampling probability of correct reasoning trajectories, but the reasoning capability itself is largely determined during pre-training.
RL does not grant the model new knowledge, nor can it compensate for missing domain expertise.

Based on these observations, we posit that long CoT symbolic reasoning training, which is highly effective for mathematical or code-generation tasks, is not directly applicable to protein function understanding.
Protein function prediction fundamentally depends on domain knowledge, structural priors, and evolutionary constraints, rather than purely abstract deduction.
The integration of large language models with external knowledge sources and computational tools has emerged as a promising approach for scientific applications requiring both reasoning and domain expertise, grounding LLM outputs in external corpora can reduce hallucination and enable knowledge-intensive question answering.
Therefore, We argue that agent-style reasoning frameworks with domain-specific tools are better aligned with the intrinsic demands of protein function understanding task.

To this end, we propose PFUA, a tool-powered protein reasoning agent that couples an online LLM with computational biology tools.
Instead of relying on unconstrained long-chain symbolic CoT, PFUA decomposes the query, invokes tools only when needed, and iteratively updates hypotheses based on verifiable tool outputs, yielding grounded reasoning traces and more reliable answers.
Extensive experiments across four benchmarks demonstrate that PFUA consistently outperforms BioMedGPT-R1, improving the average ROUGE-L recall by $98.20\%$ on Mol-Instructions.
On UniProtQA, PDB-QA, and CAFA, PFUA further surpasses BioMedGPT-R1 by $233.53\%$, $24.97\%$, and $55.57\%$, respectively.
Our main contributions are as follows:
\begin{itemize}
    \item We empirically characterize the mismatch between internal text-based reasoning and protein function understanding.
    \item we introduce PFUA, a new inference paradigm for protein function understanding that explicitly incorporates biological tools into the reasoning process.
    \item To our knowledge, we release the first multi-turn, tool-interleaved long thinking corpus for protein function understanding.
\end{itemize}

\begin{figure}[t]
  \centering
  \includegraphics[width=0.90\columnwidth]{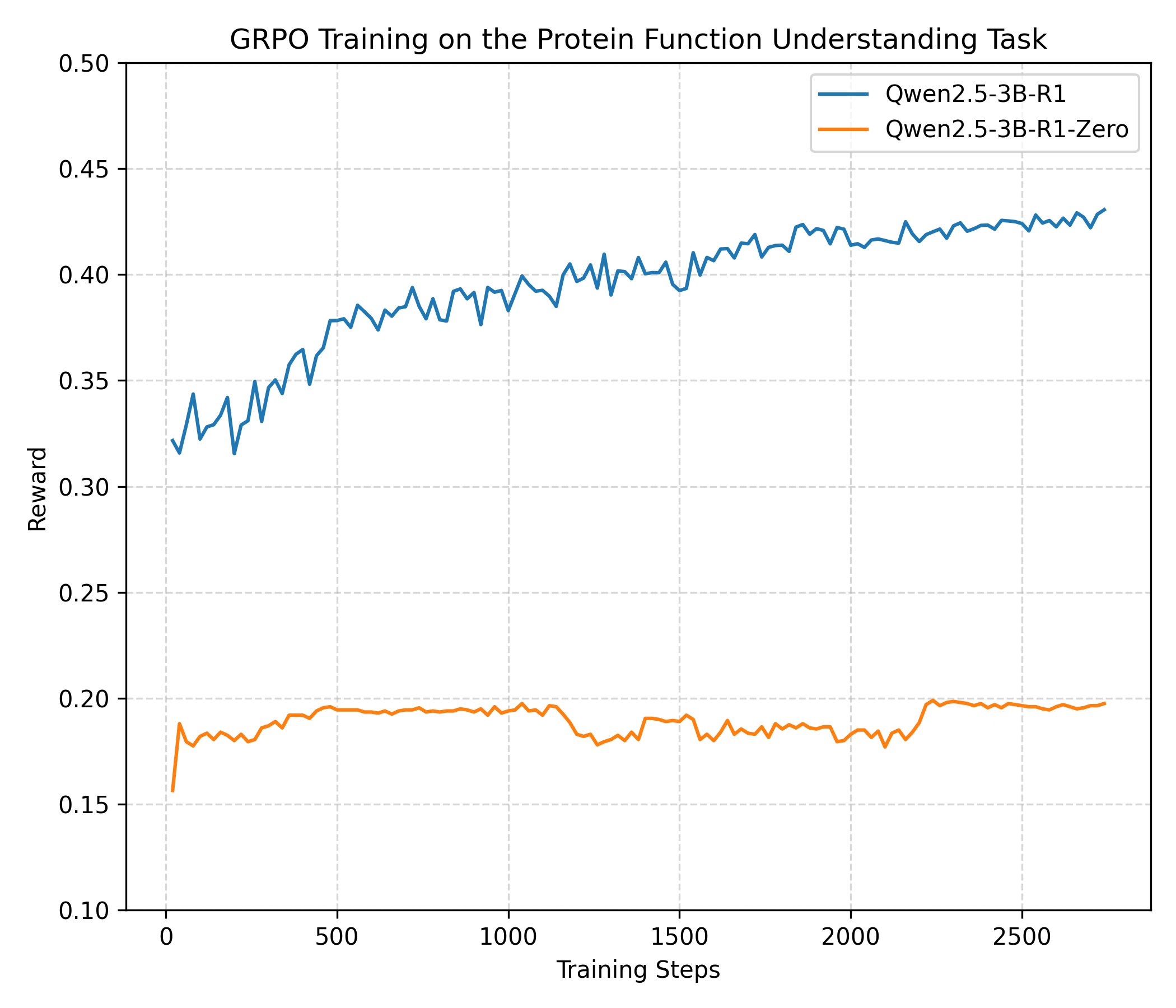}
  \caption{GRPO Training of the Protein Function Understanding Task.}
  \label{fig_2}
\end{figure}

\section{Related Work}
\subsection{LLMs for Protein Understanding}
Recent protein large language models (LLMs) reformulate protein understanding tasks into text generation problems~\cite{abdine2024prot2text,fei2025prot2text}.
These methods typically align protein sequence or structure representations with pretrained LLMs via query-based compression~\cite{liu2024prott3proteintotextgenerationtextbased}, cross-attention~\cite{Qiu2024.04.17.589642}, projection~\cite{liu2024evollamaenhancingllmsunderstanding}, or discrete tokenization strategies~\cite{ma2025prottex}.
While achieve strong empirical performance, these data-driven LLMs largely operate as black-box predictors, relying on statistical correlations rather than explicit biochemical reasoning.
The core challenge lies in interpretability and domain-specific reasoning.

\subsection{Reasoning LLMs}
Chain-of-Thought (CoT) prompting enables LLMs to perform multi-step reasoning by generating intermediate natural language explanations~\cite{wei2022chain,kojima2022large,wang2022self}.
Recent advances further improve reasoning through test-time scaling~\cite{snell2024scaling} and reinforcement learning~\cite{guo2025deepseek}, leading to strong results in domains such as mathematics~\cite{shao2024deepseekmath}.
However, in scientific applications of protein understanding, such reasoning often remains purely text-based, as exemplified by BioMedGPT-R1~\cite{luo2024biomedgpt}.
The generated rationales may reflect surface-level verbalization rather than grounded mechanistic inference, as the model lacks access to structured biological knowledge and computational validation.
This limits the applicability of standalone reasoning LLMs to complex biochemical problems.

\subsection{Tool-Powered LLMs}
The integration of large language models with external knowledge sources and computational tools has emerged as a promising approach for scientific applications requiring both reasoning and domain expertise~\cite{chen2023programthoughtspromptingdisentangling, jin2025searchr1trainingllmsreason, song2025r1searcherincentivizingsearchcapability, li2025torlscalingtoolintegratedrl}. Retrieval Augmented Generation (RAG) demonstrated how grounding LLM outputs in external corpora can reduce hallucination and enable knowledge-intensive question answering~\cite{lewis2020retrieval}. Building on this foundation, ReAct~\cite{yao2023reactsynergizingreasoningacting} interleaves reasoning with action execution, allowing models to incorporate tool outputs and observations into their reasoning process.
ReTool~\cite{feng2025retoolreinforcementlearningstrategic} proposed to leverage reinforcement learning to strategically determine when and how to invoke the code interpreter.

\section{Methods}
\begin{figure*}[t]
\centering
\includegraphics[width=0.9\textwidth]{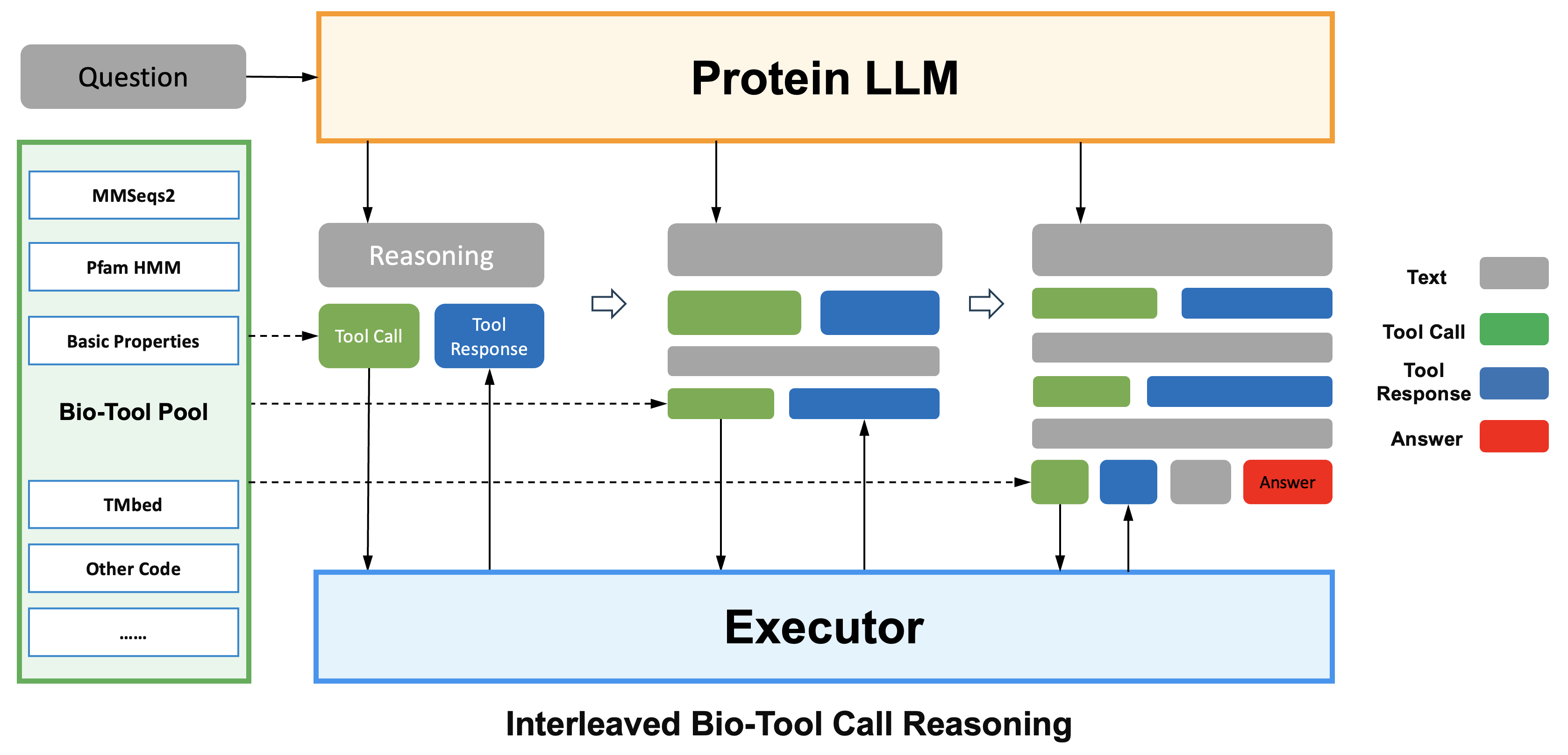}
\caption {Overview of the interleaved tool call reasoning pipeline for protein function understanding.}
\label{fig_overview}  
\end{figure*}

\subsection{Tool Pool}
In order to equip our agent with robust capabilities for autonomous protein function investigation, we construct a curated pool of computational tools~\cite{cheskis2024deciphering}.
We prioritizes tools that are programmatically accessible, provide rapid responses, and offer high evidential value for function prediction.
As shown in Figure~\ref{fig_overview}, these tools are executed through a unified executor, allowing the model to call them seamlessly within its reasoning loop.

\paragraph{Sequence basic properties.}
As a fast, mechanism-agnostic sanity check, which computes lightweight descriptors directly from the amino acid sequence.
The tool reports (i) sequence length, (ii) the maximum hydrophobic run length as a proxy for transmembrane propensity, and (iii) a low-complexity index to flag highly repetitive or compositionally biased sequences.
These features support early-stage triage: for example, extremely long hydrophobic runs suggest membrane proteins (for which soluble-enzyme assumptions may not hold), while high low-complexity scores are often associated with intrinsically disordered regions typical of regulatory proteins.
In the pipeline of catalytic activity task, for example, this tool can be used to establish a baseline hypothesis about whether a query looks enzyme-like and to prevent overconfident downstream interpretation when the sequence strongly indicates a non-enzymatic class.

\paragraph{Homology search with MMseqs2.}
To ground predictions in curated biological knowledge, we use MMseqs2 for rapid sequence similarity search against a high-quality reference database.
Specifically, we select Swiss-Prot as the target database due to its strong curation standards and rich functional annotations.
The current Swiss-Prot snapshot used in our experiments contains 573,661 entries.
Given a query sequence, this tool performs an MMseqs2 search and selects the best hit using a deterministic ranking criterion (e.g., lowest E-value, highest bit score).
It then extracts structured evidence from the corresponding Swiss-Prot record, including protein name, FUNCTION text, catalytic activity statements (reaction equations), EC numbers, cofactors, subcellular locations, and GO terms.
The resulting evidence JSON provides an auditable bridge from homology to functional inference, enabling the agent to 
(i) constrain the hypothesis space to a specific protein family/mechanism class and 
(ii) select the most appropriate catalytic reaction when multiple reactions or side activities are listed in the annotation.

\paragraph{Pfam domain analysis.}
Pfam is a widely used protein domain database that represents conserved protein families as profile hidden Markov models (HMMs).
In our tool pool, we employ Pfam HMM scanning as a primary mechanism-level analysis step.
Given a query protein sequence, we scan it against the Pfam-A HMM library to identify statistically significant domain hits, along with their alignment boundaries, coverage, and confidence scores.
Importantly, Pfam analysis constrains the functional hypothesis space at the domain and fold level before any protein-level annotation is considered.
By anchoring predictions in conserved domain families (e.g., transferase folds, oxidoreductase domains, or regulatory modules), the agent can reason about plausible biochemical mechanisms while avoiding premature commitment to overly specific functions.
Many Pfam families encompass multiple related enzymes that share a conserved fold but differ in substrate specificity or biological role. By treating Pfam hits as mechanistic and architectural evidence, rather than direct functional labels, the model avoids overinterpretation and preserves flexibility for subsequent disambiguation steps.
Subsequent tools, such as sequence homology search are then used to refine and disambiguate these domain-level hypotheses.

\paragraph{TMbed transmembrane topology prediction.}
TMbed is a protein transmembrane topology prediction tool based on large-scale protein language model embeddings.
Instead of relying on handcrafted hydrophobicity rules or shallow statistical models, TMbed leverages contextualized sequence embeddings derived from pretrained transformer models to infer transmembrane helices and membrane-associated regions with high sensitivity and robustness, particularly for short or atypical sequences.
In our tool pool, TMbed is used as a structure-aware localization discriminator that operates prior to homology-based annotation.
Given a query amino acid sequence, TMbed predicts residue-level topology states (transmembrane helix versus non-membrane) and provides a global assessment of whether the protein is likely membrane-associated.
This information is especially critical for GO annotation tasks, where cellular component (CC) terms such as membrane, endoplasmic reticulum membrane, or membrane-associated complexes fundamentally constrain the plausible functional hypotheses.
We emphasize TMbed as a high-value intermediate tool rather than a standalone annotator.
Its predictions are not interpreted in isolation; instead, they are integrated with sequence-level sanity checks and homology-based evidence to ensure consistency between predicted topology, known protein families, and curated functional annotations.
In particular, TMbed is highly informative for short proteins and small complex subunits, where traditional domain databases may provide limited coverage but membrane topology remains a decisive biological signal.

\section{Experiments}
\begin{table*}[ht]
    \centering
    \begin{tabular}{lccccc}
        \hline
        \textbf{Methods} & \textbf{Func.} & \textbf{Cat.} & \textbf{Dom.} &  \textbf{Desc.} & \textbf{Avg.} \\
        \hline
        \rowcolor{lightgray}
        \multicolumn{6}{l}{\textit{Supervised Finetuning}} \\
        BioMedGPT & 5.98 / 4.28 & 7.97 / 6.28 & 1.81 / 1.81 & 4.84 / 4.12 & 5.15 / 4.12 \\
        ProtT3 & 15.46 / 10.81  & 17.36 / 12.41 & 16.83 / 12.22 & 19.49 / 15.96 & 17.28 / 12.85 \\
        Prot2Text & 16.56 / 11.61 & 18.24 / 13.05 & 11.49 / 9.37 & 49.14 / 47.38 & 23.86 / 20.35 \\
        Qwen2.5-3B-SFT & 40.74 / 30.85 & 41.45 / 34.52 & 42.60 / 32.13 & 33.95 / 25.68 & 39.69 / 30.79 \\
        \hline
        \rowcolor{lightgray}
        \multicolumn{6}{l}{\textit{Text-based Reasoning}} \\
        BioMedGPT-R1 & 35.16 / 26.80 & 27.64 / 22.22 & 30.60 / 23.33 & 27.78 / 20.82 & 30.30 / 23.29 \\
        Qwen2.5-3B-R1 & 49.10 / 38.04 & 61.33 / 46.54 & 51.01 / 41.10 & 42.56 / 32.37 & 51.00 / 39.51 \\
        \hline
        \rowcolor{lightgray}
        \multicolumn{6}{l}{\textit{Online LLM Baseline}} \\
        DeepSeek$^{\dag}$ &  26.65 / 19/68 &  21.86 / 18.02  &  26.09 / 19.70  & 16.53 / 10.76 & 22.78 / 17.04 \\
        Kimi$^{\ddag}$ & 25.08 / 17.32  &  22.99 / 17.97  &  27.28 / 20.07  &  22.30 / 14.80  & 24.41  / 17.54 \\
        Qwen$^{\spadesuit}$ & 18.53 / 12.62  &  15.93 / 12.27  &  15.13 / 11.28  &  18.59 / 11.94  &  17.05 / 12.03 \\
        \hline
        \rowcolor{lightgray}
        \multicolumn{6}{l}{\textit{Multi-Source RAG}} \\
        DeepSeek$^{\dag}$ & 38.23 / 25.68  &  32.36 / 23.75  & 30.35  / 21.46  &  28.54 / 16.73  &  32.37 / 21.91 \\
        Kimi$^{\ddag}$ & 22.30 / 14.14  & 18.56  / 13.15  &  21.64 / 17.12  &  45.41 / 26.44  &  26.98 / 17.71 \\
        Qwen$^{\spadesuit}$ & 37.42 / 24.52  &  50.66 / 40.67  & 16.44  / 12.44  & 42.61  / 26.34  & 36.78  / 25.99 \\
        \hline
        \rowcolor{lightgray}
        \multicolumn{6}{l}{\textit{Tool-Powered Reasoning}} \\
        DeepSeek$^{\dag}$ & 59.71 / 38.31  &  47.95 / 34.15  & 48.36  / 34.36  &  58.11 / 35.70  &  53.53 / 35.63 \\
        Kimi$^{\ddag}$ & 57.98 / 35.73  &  67.68 / 46.90  &  39.76 / 28.56  &  60.08 / 36.78  &  56.38 / 36.99 \\
        Qwen$^{\spadesuit}$ (PFUA) & \textbf{66.43} / \textbf{44.29}  &  \textbf{72.32} / \textbf{54.18}  &  \textbf{54.98} / \textbf{44.26}  &  \textbf{63.60} / \textbf{41.90}  &  \textbf{64.33} / \textbf{46.16} \\
        \hline
    \end{tabular}
    \caption{Main results on protein-oriented tasks from the Mol-Instructions dataset~\cite{fang2024molinstructions}. The tasks include protein function prediction (Func.), catalytic activity prediction (Cat.), domain and motif recognition (Dom.), and general textual description generation (Desc.). For each task, performance is reported using ROUGE-1 and ROUGE-L recall (ROUGE-1 / ROUGE-L). For all online-LLM-based settings, we use $^{\dag}$\emph{DeepSeek-Reasoner}~\cite{guo2025deepseek}, $^{\ddag}$\emph{Kimi-K2-Thinking}~\cite{team2025kimi}, and $^{\spadesuit}$\emph{Qwen3-Max-Preview}~\cite{yang2025qwen3} as the backbones.}
    \label{tab:main}
\end{table*}

\subsection{Experimental Setup}
\subsubsection{Baselines}
We compare with four categories of baselines.
\textbf{(1) SFT Methods:} ProtT3~\cite{liu2024prott3proteintotextgenerationtextbased}, Prot2Text~\cite{abdine2024prot2text}, BioMedGPT~\cite{luo2024biomedgpt}, and Qwen2.5-3B-SFT~\cite{hui2024qwen2}, which rely on parametric knowledge.
\textbf{(2) Text-based reasoning (R1-style):} BioMedGPT-R1~\cite{luo2024biomedgpt} and Qwen2.5-3B-R1, which are trained to produce intermediate long reasoning traces before the final answer.
\textbf{(3) Online LLMs:} we directly prompt closed-source online LLMs to answer protein oriented queries.
\textbf{(4) Multi-source RAG:} online LLMs equipped with retrieval, where tool results from multiple sources are appended to the query as additional context.
\textbf{(5) Tool-powered protein agents:} online LLMs that interleave reasoning with explicit tool calls during inference, enabling multi-step, tool-aware decision making for protein QA.

For all online-LLM-based baselines, we choose the \emph{Kimi-K2-Thinking}~\cite{team2025kimi}, \emph{Qwen3-Max-Preview}~\cite{yang2025qwen3}, and \emph{DeepSeek-Reasoner}~\cite{guo2025deepseek} models as the backbones.
To ensure deterministic decoding, we set the sampling temperature to $0.0$ for all models (including both SFT-based and online-LLM-based baselines). 
The prompt templates for online LLM baselines, multi-source RAG, and tool-powered settings are provided in Appendix~\ref{sec_infer_temp}. 
For the Qwen2.5-3B-R1 baseline, the cold-start SFT data are synthesized using \emph{kimi-k2-0905-preview} following the template in Appendix~\ref{sec_r1_temp}.

\subsubsection{Benchmarks}
We evaluate on four protein QA benchmarks covering complementary knowledge sources and reasoning demands: Mol-Instructions (instruction-following protein/molecule tasks)~\cite{fang2024molinstructions}, UniProtQA (curated UniProt annotation-based questions on function, process, and localization)\cite{luo2024biomedgpt}, PDB-QA (queries grounded in PDB entries, emphasizing domain/topology reasoning)~\cite{liu2024prott3proteintotextgenerationtextbased}, and CAFA (GO-centric function inference in the CAFA setting)~\cite{Zhou653105}. Together, they assess instruction understanding, factual recall, and functional reasoning.

\subsubsection{Metrics}
\label{sec:metrics}
Since our models generate explicit reasoning, predictions are often longer and more descriptive than the typically concise gold references. In this case, precision-oriented metrics can be overly sensitive to additional but non-contradictory details. Therefore, we use recall-based metrics as our main results~\cite{lin2004rouge}.
ROUGE-1 Recall measures unigram coverage of the reference.
Let $R$ be the reference and $P$ the prediction. ROUGE-L Recall is computed from the longest common subsequence (LCS) and normalized by the reference length:
\begin{equation}
\mathrm{ROUGE\text{-}L} = \frac{\mathrm{LCS}(R,P)}{|R|}.
\end{equation}
These recall-based metrics emphasize whether the prediction covers the key information in the gold reference, which better matches reasoning-enhanced generation.

\begin{table*}[t]
\centering
\begin{tabular}{p{\textwidth}}
\hline
\begin{minipage}[t]{\textwidth}
\small
\textcolor{blue}{\textbf{[Question]}}\\
Analyze the protein with the following sequence and describe its properties:\\
PLAKKHNVKILPADSEHSAIFQCIQGLPEGALRRIILTASGGAFRDLPVEKLKEVKVADALKHPNWNMGKKITVDSATLFN\\
KGLEVIEAHYLFGAEYDDIEIVIHPQSIIHSMVETQDSSVLAQLGWPDMRLPILYTLSWPERVYCSEITWPRLDLCNVDLTF\\
KKPDHVKYPSMDLAYAAGRAGGTMTGVLSAANEKAVEMFIDEKISYLD
\end{minipage}
\\

\begin{minipage}[t]{\textwidth}
\small
\textcolor{blue}{\textbf{[Inference]}}\\
To analyze the given protein sequence\\
...\\
The segment `KK...K...K...R...R...KK` near C-terminus (especially `TFKK`) and central `K...R...R` suggests **bipartite or monopartite NLS**.$\backslash$ n$\backslash$ n  → Strong indicator of **nuclear localization**.$\backslash$ n$\backslash$ n- **DNA-binding motifs?** Basic region + potential helix → could resemble a **basic helix-loop-helix (bHLH)** or **zinc finger**, but only 2 Cys — not enough for classical Zn fingers.$\backslash$ n$\backslash$ n- \textcolor{red}{BLAST/Pfam homology \textbf{(simulated)}: This sequence closely resembles **human High Mobility Group proteins**}, particularly **HMGB1** or **HMGB2**.$\backslash$ n$\backslash$ n  HMGB1 signature:$\backslash$ n  - Two HMG-box domains (DNA-binding)$\backslash$ n  - Acidic C-terminal tail$\backslash$ n  - Nuclear localization$\backslash$ n  - Sequence length 200–240 aa$\backslash$ n  - High in K, A, P, E$\backslash$ n$\backslash$ n  Our sequence matches this profile.\\
...\\
**Function**: DNA bending/chromatin remodeling; pro-inflammatory DAMP when extracellular$\backslash$ n- **Homology**: Strong similarity to **HMGB1** protein$\backslash$ n- **Key Features**:$\backslash$ n  - High lysine/arginine content → DNA binding \& NLS$\backslash$ n  - Acidic patch (e.g., `EYDDIE`) → characteristic of HMGB acidic tail$\backslash$ n  - Redox-sensitive cysteines$\backslash$ n  - No signal peptide → non-classical secretion$\backslash$ n$\backslash$ n---$\backslash$ n$\backslash$ n\textcolor{red}{<answer>HMGB1-like protein involved in DNA bending and inflammatory signaling</answer>}
\end{minipage}
\\

\begin{minipage}[t]{\textwidth}
\small
\textcolor{blue}{\textbf{[Ground Truth]}}\\
A brief overview of the protein with the provided amino acid sequence is as follows: Enzyme of the plastid non-mevalonate pathway for isoprenoid biosynthesis that catalyzes the NADPH-dependent rearrangement and reduction of 1-deoxy-D-xylulose-5-phosphate (DXP) to 2-C-methyl-D- erythritol 4-phosphate (MEP). Required for chloroplast development.\\
\end{minipage}
\\
\hline
\caption{Illustration of online LLM (\emph{Qwen3-Max-Preview}) fabricating tool-call traces and purported tool outputs without actual tool execution during long chain-of-thought reasoning on Mol-Instruction (UniProtKB accession \textit{A0A1U9X9N3}).}
\label{tab:online_llm_fabricate}
\end{tabular}
\end{table*}

\subsection{Main Results}
Table~\ref{tab:main} summarizes the results on Mol-Instructions.
Overall, PFUA achieves the best performance across all four tasks and both ROUGE-1/ROUGE-L recall, substantially outperforming SFT, text-based reasoning, and multi-source RAG baselines.
This indicates that explicitly tool-aware reasoning is highly effective for protein-oriented QA, particularly when the thinking process requires integrating heterogeneous biological signals.

Compared with SFT models that mainly rely on parametric knowledge, R1-style text reasoning provides notable gains (ROUGE-L +28.32\% on Qwen2.5-3B backbone), suggesting that intermediate reasoning traces improve answer structuring.
However, simply appending multi-source tool outputs as context (RAG) yields uneven benefits and remains limited in tasks such as domain/motif recognition.
In contrast, PFUA delivers consistent improvements across \emph{Func.}, \emph{Cat.}, \emph{Dom.}, and \emph{Desc.}, with ROUGE-L +16.83\% against Qwen2.5-3B-R1 on average, supporting the advantage of actively interleaving reasoning with explicit auto tool calls to query and consolidate evidence during inference.
Moreover, the gains are especially pronounced on tasks that require precise mechanistic evidence (e.g., \emph{Dom.} and \emph{Func.}), where homology, domain boundaries, and topology signals must be jointly verified rather than heuristically inferred from text alone.
These results suggest that PFUA improves not only surface-form generation but also the reliability of evidence grounding by reducing uncertainty through targeted tool queries.
Finally, the strong and stable improvements across tasks imply better generalization to diverse protein QA intents, highlighting the robustness of tool-mediated reasoning under heterogeneous biological contexts.

\section{Analysis}
\subsection{Comparison of Three Inference Paradigms}
Table~\ref{tab:main} compares three inference paradigms under the same online-LLM backbones: (i) \emph{direct prompting} (Online LLM Baseline), (ii) \emph{passive evidence injection} (Multi-Source RAG, where tool outputs are appended as context), and (iii) \emph{tool-powered reasoning} (where the model interleaves reasoning with explicit tool calls and evidence updates). Two consistent trends emerge.
First, direct prompting yields uniformly low recall on all tasks, indicating that parametric knowledge alone is insufficient for protein-oriented questions that require precise functional, catalytic, and domain-level evidence.
Moreover, as shown in Table~\ref{tab:online_llm_fabricate}, we observe that online LLMs tend to produce \emph{tool-like} but unverifiable statements (e.g., fabricated domain names, invented hits, or arbitrary physicochemical properties), which inflates narrative plausibility but harms evidence faithfulness and downstream answer correctness.

\begin{table*}[ht]
\centering
\small
\setlength{\tabcolsep}{6pt}
\begin{tabular}{p{0.48\textwidth}|p{0.48\textwidth}}
\hline
\multicolumn{2}{p{\textwidth}}{\textbf{Question:}
Examine the given protein sequence and share a brief overview of its attributes:
MLKEFKEFALKGNVLDLAIAVVMGAAFNKIVTSLVTYIIMPLIGKIFGSVDFAKDWEFWGIKYGLFIQSIIDFIIVAIALFIFV
KIANTLVKKEEPEEEIEENTVLLTEIRDLLRAK} \\
\hline
\textbf{Text-based Reasoning:} Okay, I'm trying to examine the protein sequence and share an attribute. \textcolor{red}{Hmm, that looks a bit complicated, but I can try to break it down.}$\backslash$ n$\backslash$ nFirst, I'm looking at the function section. \textcolor{red}{Oh, wait, that's a lot of text. I think I can understand it.}$\backslash$ n$\backslash$ nThen there's the similarity section. Oh, that's a lot to take in. I can try to look up some of the mentioned proteins. $\backslash$ n$\backslash$ nThe pore-forming subunit of a porin which assembles as an undecamer (10 subunits form the central pore, 1 subunit forms the periplasmic part) to form a channel that is 2 nm wide in the outer membrane and 1. 5 nm wide in the inner membrane. The channel is non-specific and allows the passive diffusion of small molecules with a molecular weight below 600-800 Daltons through both membranes. $\backslash$ n$\backslash$ nThe homologous porins OmpC and OmpF are involved in the transport of a variety of molecules including water, ions, glucose, aminoacids, nucleoside monophosphates, lipopolysaccharide and antibiotics. $\backslash$ n\textcolor{red}{In addition, OmpF is involved in the recognition of the bacteriophage lambda receptor.} $\backslash$ n$\backslash$ nThe homologous porins OmpC and OmpF are involved in the transport of a variety of molecules including water, ions, glucose, aminoacids, nucleoside monophosphates, lipopolysaccharide and antibiotics. $\backslash$ n\textcolor{red}{In addition, OmpF is involved in the recognition of the bacteriophage lambda receptor.}& \textbf{Tool-powered Reasoning:} <answer>
This protein is a **large-conductance mechanosensitive channel (MscL)** from bacteria, functioning as a critical osmotic safety valve. Key attributes:

**Structural Features:**
- 117 amino acids in length
- Contains a major transmembrane domain spanning approximately residues 14-50 (predicted by TMbed)
- Belongs to the MscL protein family (Pfam domain PF01741.24) with 98\% coverage

**Biological Function:**
- \textcolor{blue}{Acts as a stretch-activated ion channel that opens in response to membrane tension during osmotic shock}
- Serves as an \textcolor{blue}{emergency release valve} to prevent cell lysis by allowing rapid efflux of solutes when cells experience sudden hypoosmotic stress
- Located in the cell membrane as a multi-pass transmembrane protein

**Key Evidence:**
- Extremely significant Pfam hit to the MscL domain (e-value: 8e-39)
- Transmembrane prediction confirms membrane integration despite initial heuristic suggesting otherwise

This protein represents a fundamental bacterial survival mechanism, allowing cells to \textcolor{blue}{rapidly respond to osmotic downshock by releasing cytoplasmic solutes} before membrane rupture occurs.
</answer> \\[0.3cm]
\hline
\multicolumn{2}{p{\textwidth}}{\textbf{Answer:} A brief overview of the protein with the provided amino acid sequence is as follows: \textcolor{blue}{Channel that opens in response to stretch forces in the membrane lipid bilayer.} May participate in the \textcolor{blue}{regulation of osmotic pressure changes within the cell}.}\\
\hline
\end{tabular}
\caption{Case study of tool-powered reasoning (PFUA) vs.\ text-only reasoning (BioMedGPT-R1) on Mol-Instruction (Task: \textit{general\_function}, UniProtKB accession: \textit{A0A1U9X9N3}). The full tool-call traces and intermediate reasoning of PFUA are in Appendix~\ref{sec_full_case}.}
\label{tab:reasoning_comparison}
\end{table*}

\subsection{More Benchmarks}

\begin{figure}[ht]
  \centering
  \includegraphics[width=0.85\columnwidth]{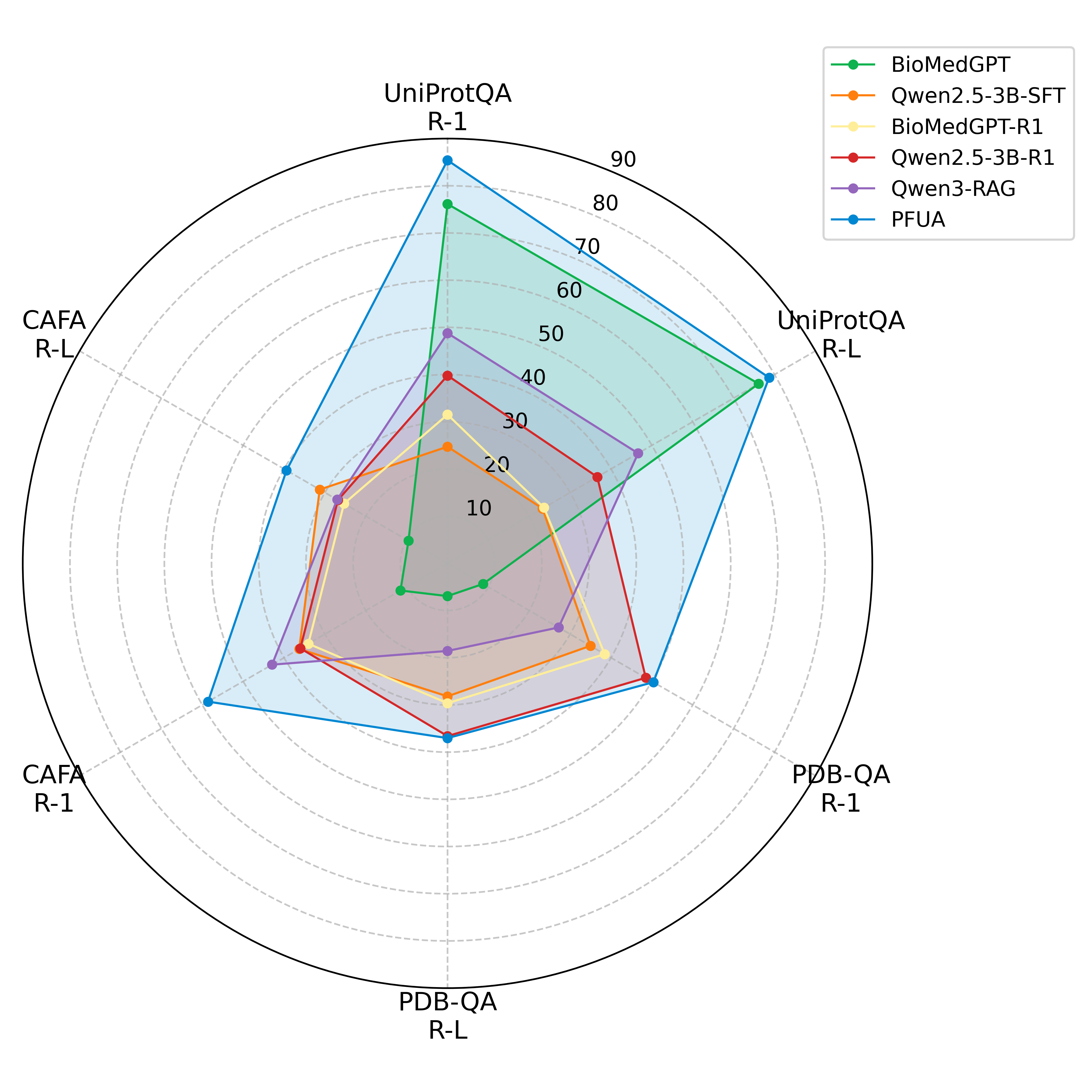}
  \caption{Results on three additional protein QA benchmarks. Performance is reported using ROUGE-1 and ROUGE-L recall (ROUGE-1 / ROUGE-L). The backbone online LLM of Qwen3-RAG and PFUA is \emph{Qwen3-Max-Preview}.}
  \label{fig_10}
\end{figure}

\paragraph{RAG helps, but remains inconsistent and brittle.}
Multi-Source RAG improves over direct prompting for some backbones (e.g., Qwen: 17.05/12.03 $\rightarrow$ 36.78/25.99 on Avg.), suggesting that providing external evidence is necessary. However, the gains are not uniform: Kimi shows only marginal improvement on Avg. (24.41/17.54 $\rightarrow$ 26.98/17.71), and per-task performance can remain unstable, especially for domain/motif recognition. This indicates a key limitation of passive context augmentation: simply concatenating heterogeneous tool outputs does not guarantee that the model will \emph{select}, \emph{prioritize}, and \emph{integrate} the right evidence, nor does it prevent partial misinterpretation of tool results.

\paragraph{Tool-powered reasoning yields robust, backbone-agnostic gains.}
In contrast, tool-powered reasoning consistently delivers large improvements across all four tasks for every backbone.
Relative to direct prompting, this corresponds to substantial overall gains on Avg. (e.g., +277\% in ROUGE-1 and +284\% in ROUGE-L on \emph{Qwen3-Max-Preview} backbone).
The consistent uplift suggests that the key factor is not merely \emph{access} to external evidence, but \emph{how} evidence is operationalized during inference: explicit tool calls enforce grounded intermediate states, enable iterative hypothesis revision, and reduce the tendency to hallucinate tool-derived facts.
Overall, these results support tool-powered protein agents as a more reliable and scalable paradigm than either parametric-only prompting or passive multi-source RAG.

Figure~\ref{fig_10} extends evaluation to UniProtQA, PDB-QA, and CAFA, which emphasize curated annotation recall and GO-centric function inference. Overall, PFUA achieves the best performance on all three benchmarks, indicating that tool-powered reasoning generalizes beyond Mol-Instructions and remains effective under diverse evidence types and question styles.
A closer look shows that Multi-Source RAG brings only moderate and sometimes unstable gains, especially on PDB-QA. In contrast, PFUA yields substantial improvements over the strongest non-tool baseline BioMedGPT on UniProtQA (+12.2\% / +3.4\% relatively in ROUGE-1/ROUGE-L).

\subsection{Case Study}

Table~\ref{tab:reasoning_comparison} presents a representative example from Mol-Instruction (\textit{general\_function}) to compare text-only reasoning (BioMedGPT-R1) with tool-powered reasoning (PFUA). Given the sequence, BioMedGPT-R1 generates a long narrative that explicitly references ``function'' and ``similarity'' sections and then copies a porin-centric description (e.g., OmpC/OmpF, bacteriophage lambda receptor), which is not supported by any sequence-derived evidence. The reasoning also contains self-referential, low-information statements (highlighted in red) and repeated fragments, indicating that the model is hallucinating a UniProt-like annotation page and filling it with plausible-looking content rather than inferring attributes from the input sequence.

In contrast, PFUA produces an answer that matches the expected label by grounding the summary in tool evidence (full tool-call traces in Appendix~\ref{sec_full_case}). The tool-powered summary correctly identifies the protein as a bacterial mechanosensitive channel (MscL) and states the key functional property required by the task---a stretch-activated membrane channel involved in osmotic downshock protection. Importantly, PFUA’s final answer is concise and aligned with the sequence-level signals (membrane protein with characteristic MscL family/domain support), whereas the text-only baseline drifts to an unrelated porin narrative despite its lengthy chain-of-thought.

\section{Conclusion}
In this work, we investigated the limitations of purely text-based reasoning paradigms for protein function understanding and showed that R1-style chain-of-thought reinforcement learning fails to deliver substantial gains in this knowledge-intensive scientific task. Our analyses demonstrate that protein function prediction relies critically on external biological priors and verifiable computational evidence, which cannot be acquired through internal reasoning alone. To bridge this gap, we proposed PFUA, a tool-augmented protein reasoning agent that integrates large language models with domain-specific biological tools, enabling grounded intermediate evidence and more reliable predictions. Extensive experiments across multiple benchmarks confirm that PFUA significantly outperforms text-only reasoning models while providing interpretable reasoning traces. This work highlights tool-integrated agents as a promising paradigm for scientific AI systems, and we anticipate that future research will extend this framework to broader bioinformatics tasks and richer tool ecosystems.

\section*{Limitations}
First, our study mainly demonstrates the effectiveness of tool-augmented reasoning under a fixed tool pool; the design and optimization of the tool set itself are not explored and may further affect performance.
Second, although the datasets are constructed following established protocols, certain samples may still require additional manual verification to reduce potential annotation noise.
Third, our experiments focus on protein QA, and the effectiveness of tool augmentation on broader protein-related tasks remains to be systematically validated, such as enzyme optimization, protein design, protein–protein interaction prediction, and molecular docking.
Finally, the current evaluation relies on concise gold answers, which may not fully reflect the quality of more elaborate reasoning traces; designing metrics that better align concise references with complex reasoning outputs is left for future work.

\section*{Acknowledgments}
This work was supported by the National Natural Science Foundation of China (No. 62476187), the National Science and Technology Major Project (No. 2022ZD0115002), and Project Funded by the Priority Academic Program Development of Jiangsu Higher Education Institutions.

\bibliography{custom}

\clearpage
\appendix

\section{Templates of Three Inference Paradigms}
\label{sec_infer_temp}

\begin{table*}[ht]
\centering
\renewcommand{\arraystretch}{1.15}
\begin{tabular}{p{\textwidth}}
\hline
\rowcolor{lightgray}
\textbf{Template for Online LLM Baseline}\\
\begin{minipage}[t]{\textwidth}
\small
\textbf{[ROLE]}\\
You are a professional bioinformatics assistant.\\

\textbf{[TASK]}\\
Please first provide detailed reasoning and analysis.\\

\textbf{[CONSTRAINTS]}\\
Then give a concise final answer wrapped strictly inside <answer></answer> tags.\\
\end{minipage}
\\
\hline
\rowcolor{lightgray}
\textbf{Template for Multi-Source RAG} \\
\begin{minipage}[t]{\textwidth}
\small
\textbf{[ROLE]}\\
You are an expert protein analysis assistant.\\

\textbf{[TASK]}\\
Analyze the given protein sequence. You are provided with external tool outputs (computed properties, homology search, domain scan, and topology prediction).\\
Use these tool results as evidence to reasoning and answer the question.\\

\textbf{[CONSTRAINTS]}\\
- Do NOT request additional tools or external calls. Everything you need is already included below.\\
- The final answer MUST be wrapped in <answer>...</answer>.\\
\end{minipage}
\\
\hline
\rowcolor{lightgray}
\textbf{Template for Tool-Powered Reasoning} \\
\begin{minipage}[t]{\textwidth}
\small
\textbf{[ROLE]}\\
You are an expert protein analysis agent.\\

\textbf{[TASK]}\\
Your goal is to analyze the protein sequence and produce a biologically meaningful interpretation.\\
You should reason step-by-step, form hypotheses, and use tools only when they help reduce uncertainty.\\

\textbf{[REASONING REQUIREMENTS]} \\
Before calling tools, you MUST:\\
- propose hypotheses about the protein\\
- explain which uncertainties still remain\\
For EVERY tool call, you MUST:\\
- explicitly explain WHY this tool is needed\\
- describe WHAT evidence you expect it to provide\\
After each tool result, you MUST:\\
- summarize what new evidence was obtained\\
- update or revise your hypothesis\\
- decide whether additional tools are needed\\

\textbf{[TOOLS]} \\
You may call the following tools through function calling:\\
- $seq\_basic\_props$: basic physicochemical properties\\
- $pfam\_hmmscan$: domain and family inference\\
- $mmseqs2\_besthit\_uniprot$: homolog search and functional annotation\\
- $tmbed\_predict$: transmembrane and topology prediction\\
Prefer \texttt{{"sequence\_ref": "query"}} instead of pasting long sequences.\\

\textbf{[OUTPUT]}\\
When finished, wrap the final answer in <answer>...</answer>.
\end{minipage}
\\
\hline
\caption{Inference templates for online LLM, multi-source RAG and tool-powered protein reasoning.}
\label{tab:inference_template}
\end{tabular}
\end{table*}

Table~\ref{tab:inference_template} summarizes the prompt templates used to instantiate three representative inference paradigms for protein understanding. The \emph{online LLM baseline} adopts a minimal instruction-only setting, where the model relies solely on internal parametric knowledge to produce a step-by-step analysis followed by a concise answer. The \emph{multi-source RAG} template augments the model with pre-collected evidence from heterogeneous tools (e.g., computed properties, homology search, domain scanning, and topology prediction), and explicitly constrains the model to ground its reasoning in the provided outputs without requesting any additional external calls. In contrast, the \emph{tool-powered reasoning} template frames the model as an agent that can actively decide \emph{when} and \emph{why} to invoke specific bioinformatics tools, requiring hypothesis-driven reasoning, uncertainty tracking, and iterative belief updates after each tool result. Collectively, these templates establish a controlled comparison from static, tool-free inference, to evidence-conditioned RAG, and finally to adaptive, decision-centric tool use for more reliable and interpretable protein analysis.

\section{Case Study of Full Tool-Call Reasoning}
\label{sec_full_case}
As shown in Table~\ref{tab:inference_template1},~\ref{tab:inference_template2},~\ref{tab:inference_template3},
to provide a concrete and fully transparent view of how our tool-powered paradigm operates in practice, we include a complete case study that records the agent’s reasoning trajectory, including all intermediate hypotheses, tool-invocation decisions, and evidence-driven updates.
This example is intentionally placed in the appendix because it is substantially longer than typical main-text examples, but it serves as an important qualitative supplement to the quantitative results.
Specifically, the case study demonstrates how the agent (i) starts from sequence-level cues to form initial functional hypotheses, (ii) identifies key uncertainties that cannot be resolved reliably from parametric knowledge alone, (iii) selects appropriate tools (e.g., basic physicochemical profiling, domain scanning, homology search, and topology prediction) with explicit expectations of the evidence each tool should provide, and (iv) iteratively revises its interpretation after observing tool outputs.
Overall, the full trace illustrates the core advantage of tool-call reasoning: rather than producing a single-shot explanation, the agent performs hypothesis-driven evidence acquisition and belief updating, yielding a more grounded and auditable protein interpretation.

\begin{table*}[ht]
\centering
\renewcommand{\arraystretch}{1.15}
\begin{tabular}{p{\textwidth}}
\hline
\begin{minipage}[t]{\textwidth}
\small
\textcolor{blue}{\textcolor{blue}{<|im\_start|>user}}\\
\textbf{[ROLE]}\\
You are an expert protein analysis agent.\\
\textbf{[TASK]}\\
Your goal is to analyze the protein sequence and produce a biologically meaningful interpretation.\\
You should reason step-by-step, form hypotheses, and use tools only when they help reduce uncertainty.\\
\textbf{[REASONING REQUIREMENTS]} \\
Before calling tools, you MUST:\\
- propose hypotheses about the protein\\
- explain which uncertainties still remain\\
For EVERY tool call, you MUST:\\
- explicitly explain WHY this tool is needed\\
- describe WHAT evidence you expect it to provide\\
After each tool result, you MUST:\\
- summarize what new evidence was obtained\\
- update or revise your hypothesis\\
- decide whether additional tools are needed\\
\textbf{[TOOLS]} \\
You may call the following tools through function calling:\\
- $seq\_basic\_props$: basic physicochemical properties\\
- $pfam\_hmmscan$: domain and family inference\\
- $mmseqs2\_besthit\_uniprot$: homolog search and functional annotation\\
- $tmbed\_predict$: transmembrane and topology prediction\\
Prefer \texttt{{"sequence\_ref": "query"}} instead of pasting long sequences.\\
\textbf{[OUTPUT]}\\
When finished, wrap the final answer in <answer>...</answer>.\\
Examine the given protein sequence and share a brief overview of its attributes:\\
Protein sequence (id=query):\\
```MLKEFKEFALKGNVLDLAIAVVMGAAFNKIVTSLVTYIIMPLIGKIFGSVDFAKDWEFWGIKYGLFIQSIIDFIIVAIAL\\FIFVKIANTLVKKEEPEEEIEENTVLLTEIRDLLRAK```\\
\textcolor{blue}{<|im\_end|>}\\

\textcolor{blue}{<|im\_start|>assistant}\\
I need to analyze this protein sequence step by step. Let me start by examining the sequence and forming initial hypotheses.

The sequence is:\\
MLKEFKEFALKGNVLDLAIAVVMGAAFNKIVTSLVTYIIMPLIGKIFGSVDFAKDWEFWGIKYGLFIQSIIDFIIVAIAL\\FIFVKIANTLVKKEEPEEEIEENTVLLTEIRDLLRAK

Initial observations:
1. The sequence starts with "MLKEFKEFALK" - this looks like it could be a signal peptide or N-terminal region
2. There are several hydrophobic regions (like "LAIAVVMGAAFNKIVTSLVTYIIMPLIGKIFG")
3. There are charged regions with many glutamates (E) and lysines (K) at the end
4. The length appears to be around 120-130 amino acids

Hypotheses:
1. This could be a membrane protein due to the hydrophobic stretches
2. It might have transmembrane domains
3. The C-terminal region with many charged residues might be cytoplasmic
4. It could be involved in transport or signaling based on the domain structure

Uncertainties that remain:
1. Exact length and composition
2. Whether it has transmembrane domains and how many
3. What protein family it belongs to
4. Its specific function
5. Its subcellular localization

I should start with basic properties to get fundamental information about the sequence. This will help me understand the basic characteristics before diving into more specific analyses.

Let me call seq\_basic\_props first to get the basic properties.\\
\textcolor{blue}{<tool\_call>}\\
\{"name": "seq\_basic\_props", "arguments": \{"sequence\_ref": "query"\}\}\\
\textcolor{blue}{</tool\_call>}\\\textcolor{blue}{<|im\_end|>}\\

\textcolor{blue}{<|im\_start|>user}\\
\textcolor{blue}{<tool\_response>}\\
\{"length": 117, "hydrophobic\_run\_max": 12, "low\_complexity\_index\_0to1": 0.1171, "heuristics": \{"looks\_membrane\_like": false, "looks\_low\_complexity\_like": false\}\}\\
\textcolor{blue}{</tool\_response>}\\\textcolor{blue}{<|im\_end|>}\\
\end{minipage}\\
\hline
\caption{Full reasoning traces of PFUA on Mol-Instruction (Task: \textit{general\_function}, UniProtKB accession: \textit{A0A1U9X9N3}) Part 1.}
\label{tab:inference_template1}
\end{tabular}
\end{table*}

\begin{table*}[ht]
\centering
\renewcommand{\arraystretch}{1.15}
\begin{tabular}{p{\textwidth}}
\hline
\begin{minipage}[t]{\textwidth}
\small
\textcolor{blue}{<|im\_start|>assistant}\\
Now I have the basic properties:
- Length: 117 amino acids
- Maximum hydrophobic run: 12 residues
- Low complexity index: 0.1171 (relatively low)
- Heuristics suggest it doesn't look membrane-like or low-complexity-like

This is interesting because my initial hypothesis about it being membrane-like might be incorrect based on the heuristic. However, I should still check for transmembrane domains more thoroughly since the hydrophobic run of 12 residues could still indicate potential transmembrane regions.

Let me now search for homologs in UniProt to get functional annotation and see what similar proteins are known to do. This will help me understand the protein's function and validate or revise my hypotheses.\\
\textcolor{blue}{<tool\_call>}\\
\{"name": "mmseqs2\_besthit\_uniprot", "arguments": \{"sequence\_ref": "query", "min\_seq\_id": 0.3\}\}\\
\textcolor{blue}{</tool\_call>}\\\textcolor{blue}{<|im\_end|>}\\

\textcolor{blue}{<|im\_start|>user}\\
\textcolor{blue}{<tool\_response>}\\
\{"best\_hit": \{"query": "query", "target": "Q4L656", "pident": 100.0, "alnlen": 117, "evalue": 1.299e-66, "bits": 222.0\}, "uniprot\_annotation": \{"accessions": ["Q4L656"], "protein\_name": "Large-conductance mechanosensitive channel {ECO:0000255|HAMAP-Rule:MF\_00115}", "function": ["Channel that opens in response to stretch forces in the", "membrane lipid bilayer. May participate in the regulation of osmotic", "pressure changes within the cell. {ECO:0000255|HAMAP-Rule:MF\_00115}."], "catalytic\_activity": [], "ec": [], "cofactor": [], "subcellular\_location": ["Cell membrane {ECO:0000255|HAMAP-Rule:MF\_00115};", "Multi-pass membrane protein {ECO:0000255|HAMAP-Rule:MF\_00115}."], "go": ["DR   GO; GO:0005886; C:plasma membrane; IEA:UniProtKB-SubCell.", "DR   GO; GO:0008381; F:mechanosensitive monoatomic ion channel activity; IEA:UniProtKB-UniRule."]\}\}\\
\textcolor{blue}{</tool\_response>}\\\textcolor{blue}{<|im\_end|>}\\

\textcolor{blue}{<|im\_start|>assistant}\\
Great! The UniProt search revealed that this is a large-conductance mechanosensitive channel protein (MscL). This is very informative. The annotation shows:

1. It's a channel that opens in response to stretch forces in the membrane lipid bilayer
2. It participates in regulation of osmotic pressure changes within the cell
3. It's located in the cell membrane and is a multi-pass membrane protein
4. It has mechanosensitive monoatomic ion channel activity

This contradicts the initial heuristic from seq\_basic\_props that said it doesn't look membrane-like. This makes sense because mechanosensitive channels are indeed membrane proteins. The heuristic might not have been sensitive enough to detect the transmembrane nature.

Now I should check for transmembrane domains to confirm this and understand the topology. The annotation says it's a multi-pass membrane protein, so I expect multiple transmembrane regions.

Let me use tmbed\_predict to get the transmembrane topology prediction.

Excellent! The UniProt search revealed that this is a large-conductance mechanosensitive channel protein (MscL). This is very informative. The annotation shows:

1. It's a channel that opens in response to stretch forces in the membrane lipid bilayer
2. It participates in regulation of osmotic pressure changes within the cell
3. It's located in the cell membrane and is a multi-pass membrane protein
4. It has mechanosensitive monoatomic ion channel activity

This contradicts the initial heuristic from seq\_basic\_props that said it doesn't look membrane-like. This makes sense because mechanosensitive channels are indeed membrane proteins. The heuristic might not have been sensitive enough to detect the transmembrane nature.

Now I should check for transmembrane domains to confirm this and understand the topology. The annotation says it's a multi-pass membrane protein, so I expect multiple transmembrane regions.

Let me use tmbed\_predict to get the transmembrane topology prediction.\\
\textcolor{blue}{<tool\_call>}\\
\{"name": "tmbed\_predict", "arguments": \{"sequence\_ref": "query"\}\}\\
\textcolor{blue}{</tool\_call>}\\\textcolor{blue}{<|im\_end|>}\\

\textcolor{blue}{<|im\_start|>user}\\
\textcolor{blue}{<tool\_response>}\\
\{"prediction": \{"raw\_pred": ">query $\backslash$ n MLKEFKEFALKGNVLDLAIAVVMGAAFNKIVTSLVTYIIMPLIGKIFGSVDFAKDWEFWGIKYGLFIQSIIDFIIVAIALFIFVKIANTLVKKEEPEEEIEENTVLLTEIRDLLRAK $\backslash$ n.............HHHHHHHHHHHHHHHHHHHHHHHHHHHHHHHHH.................hhhhhhhhhhhhhhhhhhhhhhhh..............................", "tm\_signal\_letter\_hits": 36, "has\_tm\_signal\_heuristic": true\}\}\\
\textcolor{blue}{</tool\_response>}\\\textcolor{blue}{<|im\_end|>}\\
\end{minipage}\\
\hline
\caption{Full reasoning traces of PFUA on Mol-Instruction (Task: \textit{general\_function}, UniProtKB accession: \textit{A0A1U9X9N3}) Part 2.}
\label{tab:inference_template2}
\end{tabular}
\end{table*}

\begin{table*}[ht]
\centering
\renewcommand{\arraystretch}{1.15}
\begin{tabular}{p{\textwidth}}
\hline
\begin{minipage}[t]{\textwidth}
\small
\textcolor{blue}{<|im\_start|>assistant}\\
Now I have the transmembrane prediction results. Let me analyze what this shows:

The raw prediction shows:
- A long stretch of "H" characters (representing helical transmembrane regions) starting around position 14 and continuing for about 36 residues
- Then a stretch of "h" characters (likely representing another type of transmembrane or membrane-associated region) later in the sequence
- The prediction indicates 36 tm\_signal\_letter\_hits and has\_tm\_signal\_heuristic: true

This confirms that this is indeed a transmembrane protein, which aligns with the UniProt annotation that it's a "Multi-pass membrane protein". The initial heuristic from seq\_basic\_props that said "looks\_membrane\_like: false" was incorrect - this is definitely a membrane protein.

Let me also check for Pfam domains to get more detailed information about the domain architecture and confirm this is indeed an MscL channel.\\
\textcolor{blue}{<tool\_call>}\\
\{"name": "pfam\_hmmscan", "arguments": \{"sequence\_ref": "query"\}\}\\
\textcolor{blue}{</tool\_call>}\\\textcolor{blue}{<|im\_end|>}\\

\textcolor{blue}{<|im\_start|>user}\\
\textcolor{blue}{<tool\_response>}\\
\{"hits": [\{"pfam\_id": "MscL", "pfam\_acc": "PF01741.24", "query": "query", "evalue": 8e-39, "score": 133.3, "hmm\_from": 1, "hmm\_to": 124, "ali\_from": 1, "ali\_to": 115, "coverage\_query": 0.9829, "desc": "Large-conductance mechanosensitive channel, MscL"\}, \{"pfam\_id": "Anoctamin", "pfam\_acc": "PF04547.18", "query": "query", "evalue": 0.045, "score": 13.1, "hmm\_from": 129, "hmm\_to": 199, "ali\_from": 23, "ali\_to": 102, "coverage\_query": 0.6838, "desc": "Calcium-activated chloride channel"\}, \{"pfam\_id": "Mim2", "pfam\_acc": "PF19117.6", "query": "query", "evalue": 0.047, "score": 13.8, "hmm\_from": 15, "hmm\_to": 31, "ali\_from": 32, "ali\_to": 48, "coverage\_query": 0.1453, "desc": "Mitochondrial import 2"\}, \{"pfam\_id": "UPF0239", "pfam\_acc": "PF06783.17", "query": "query", "evalue": 0.16, "score": 12.8, "hmm\_from": 21, "hmm\_to": 60, "ali\_from": 61, "ali\_to": 99, "coverage\_query": 0.3333, "desc": "Uncharacterised protein family (UPF0239)"\}, \{"pfam\_id": "DUF3917", "pfam\_acc": "PF13055.11", "query": "query", "evalue": 0.32, "score": 12.0, "hmm\_from": 21, "hmm\_to": 68, "ali\_from": 33, "ali\_to": 84, "coverage\_query": 0.4444, "desc": "Protein of unknown function (DUF3917)"\}], "selected\_domains": [\{"pfam\_id": "MscL", "pfam\_acc": "PF01741.24", "query": "query", "evalue": 8e-39, "score": 133.3, "hmm\_from": 1, "hmm\_to": 124, "ali\_from": 1, "ali\_to": 115, "coverage\_query": 0.9829, "desc": "Large-conductance mechanosensitive channel, MscL"\}]\}\\
\textcolor{blue}{</tool\_response>}\\\textcolor{blue}{<|im\_end|>}\\

\textcolor{blue}{<|im\_start|>assistant}\\
Now I have all the key information to provide a comprehensive analysis. Let me synthesize the findings:

1. Basic properties:
   - Length: 117 amino acids
   - Maximum hydrophobic run: 12 residues
   - Low complexity index: 0.1171 (relatively low)
   - Initial heuristic suggested it didn't look membrane-like, but this was incorrect

2. UniProt homology search:
   - 100\% identity to Q4L656 (Large-conductance mechanosensitive channel)
   - Function: Channel that opens in response to stretch forces in membrane lipid bilayer
   - Role: Regulation of osmotic pressure changes within the cell
   - Subcellular location: Cell membrane, multi-pass membrane protein
   - GO terms: plasma membrane, mechanosensitive monoatomic ion channel activity

3. Transmembrane prediction:
   - Confirmed transmembrane signal with 36 hits
   - Shows a long hydrophobic transmembrane region (the HHHHH... stretch)
   - Confirms it's a membrane protein despite initial heuristic

4. Pfam domain analysis:
   - Strong hit to MscL domain (PF01741.24) with extremely low e-value (8e-39)
   - Coverage of 98.29\% of the query sequence
   - Other weaker hits are likely false positives or overlapping domains
   - The MscL domain is definitive for large-conductance mechanosensitive channels

This is clearly a bacterial mechanosensitive channel protein that acts as an emergency release valve during osmotic shock. The transmembrane prediction shows the characteristic hydrophobic region that spans the membrane, and the Pfam analysis confirms it belongs to the MscL family.

The initial heuristic that said it didn't look membrane-like was incorrect - this is a classic membrane protein. The hydrophobic run of 12 residues was actually part of a larger transmembrane domain.

I can now provide a comprehensive final answer.

\textcolor{red}{<answer>}\\
This protein is a **large-conductance mechanosensitive channel (MscL)** from bacteria, functioning as a critical osmotic safety valve. Key attributes:

**Structural Features:**
- 117 amino acids in length
- Contains a major transmembrane domain spanning approximately residues 14-50 (predicted by TMbed)
- Belongs to the MscL protein family (Pfam domain PF01741.24) with 98

**Biological Function:**
- Acts as a stretch-activated ion channel that opens in response to membrane tension during osmotic shock
- Serves as an emergency release valve to prevent cell lysis by allowing rapid efflux of solutes when cells experience sudden hypoosmotic stress
- Located in the cell membrane as a multi-pass transmembrane protein

**Key Evidence:**
- Extremely significant Pfam hit to the MscL domain (e-value: 8e-39)
- Transmembrane prediction confirms membrane integration despite initial heuristic suggesting otherwise

This protein represents a fundamental bacterial survival mechanism, allowing cells to rapidly respond to osmotic downshock by releasing cytoplasmic solutes before membrane rupture occurs.\\
\textcolor{red}{</answer>}\\
\textcolor{blue}{<|im\_end|>}\\
\end{minipage}\\
\hline
\caption{Full reasoning traces of PFUA on Mol-Instruction (Task: \textit{general\_function}, UniProtKB accession: \textit{A0A1U9X9N3}) Part 3.}
\label{tab:inference_template3}
\end{tabular}
\end{table*}

\section{Synthesizing R1-Style Reasoning Traces}
\label{sec_r1_temp}
For the baseline Qwen2.5-3B-R1 model, which adopts a standard DeepSeek-R1-style training pipeline, 
we construct synthetic reasoning traces as cold-start supervision. 
Specifically, we prompt an online LLM, Kimi (\emph{kimi-k2-0905-preview}), with the original Mol-Instruction inputs 
(i.e., the question and the protein sequence) using the template in Table~\ref{tab:r1_template}. 
This procedure yields 2{,}000 synthetic cold-start SFT examples for training Qwen2.5-3B-R1.

\begin{table}[ht]
\small
\centering
\renewcommand{\arraystretch}{1.15}
\setlength{\tabcolsep}{8pt}
\begin{tabular}{p{0.98\linewidth}}
\hline
\rowcolor{lightgray}
\textbf{Template for Synthesizing Reasoning Trace} \\
You are an expert AI in bioinformatics and computational biology. Your task is to 
generate a detailed, step-by-step intermediate reasoning process that connects the 
given input to the given output. 
The reasoning process should explain how one might logically derive the output 
from the input.

\textbf{Input:}\\
question: \texttt{``{question}"}\\
protein sequence: \texttt{``{sequence}"}\\

\textbf{Output:}\\
answer: \texttt{``{answer}''}

Your response should consist solely of the reasoning process enclosed in 
\texttt{<think>} ... \texttt{</think>}. \\
\hline
\end{tabular}
\caption{Cold-start template for synthesizing reasoning traces used to train Qwen2.5-3B-R1 (generated via \emph{kimi-k2-0905-preview}).}
\label{tab:r1_template}
\end{table}

\end{document}